\documentclass{article}

\usepackage[dblblindworkshop, final]{neurips_2026}

\usepackage[utf8]{inputenc}
\usepackage[T1]{fontenc}
\usepackage{hyperref}
\usepackage{url}
\usepackage{booktabs}
\usepackage{amsmath,amssymb,amsfonts}
\usepackage{nicefrac}
\usepackage{microtype}
\usepackage[table]{xcolor}
\usepackage{graphicx}
\usepackage{textcomp}
\usepackage{bm}
\usepackage{bbm}
\usepackage{subcaption}
\usepackage{siunitx}

\title{Support-Constrained Offline Policy Improvement for Pediatric ECMO from Limited Clinical Data}
\workshoptitle{World Models for High-Stakes Health}

\author{
Fateme Golivand Darvishvand \\
University of Texas at Dallas\\
\texttt{fateme.golivanddarvishvand@utdallas.edu}\\
\And
Saurabh Mathur\\
Technische Universit\"at Darmstadt\\
\texttt{saurabh.mathur@tu-darmstadt.de}\\
\And
Nikhilesh Prabhakar\\
University of Texas at Dallas\\
\texttt{Nikhilesh.Prabhakar@UTDallas.edu}\\
\And
Michael A. Skinner\\
University of Texas at Dallas\\
\texttt{mas140130@utdallas.edu}\\
\And
Ameet Soni\\
Swarthmore College\\
\texttt{asoni1@swarthmore.edu}\\
\And
Neel Shah\\
Washington University in St. Louis\\
\texttt{neel.shah@wustl.edu}\\
\And
Ethan L Sanford\\
UT Southwestern Medical Center\\
\texttt{Ethan.Sanford@UTSouthwestern.edu}\\
\And
Lakshmi Raman\\
UT Southwestern Medical Center\\
\texttt{Lakshmi.Raman@UTSouthwestern.edu}\\
\And
Sriraam Natarajan\\
University of Texas at Dallas\\
\texttt{Sriraam.Natarajan@utdallas.edu}\\
}

\begin{document}

\maketitle

\begin{abstract}
Managing children on ECMO (Extracorporeal Membrane Oxygenation) requires balancing short-term physiological stability with the long-term goal of preventing neurological injury. Standard reinforcement learning can propose risky, out-of-distribution actions, while imitation learning is limited to reproducing clinician behavior. We introduce Support-Constrained Fitted Q-Iteration (FQI-SC), an offline reinforcement learning framework that cautiously improves upon expert behavior by restricting optimization to empirically supported state-action pairs and defaulting to the clinician baseline in low-data regions. On 78 pediatric trajectories totaling over 6,000 hours of clinical data, we evaluate our framework using Fitted Q-Evaluation (FQE) and a behavioral disagreement. Additionally, we formalize the performance-safety trade-off by introducing a unified Balance Score ($S(\pi)$). Our empirical evaluation shows that support-constrained policy improves the Balance Score by 38\% relative to FQI, improves estimated value by 1.28 FQE units over imitation learning, and reduces disagreement with clinician actions by 71\% relative to FQI.
\end{abstract}

\section{Introduction}

Extracorporeal Membrane Oxygenation (ECMO) is a life-sustaining intervention for respiratory or cardiac failure~\citep{ecmo}. ECMO management requires continuous circuit adjustment to maintain physiological stability while minimizing severe complications such as neurological injury. This real-time decision-making is especially difficult in pediatric cases.

Offline Reinforcement Learning (RL) offers a framework for optimizing treatment from historical trajectories~\citep{levine2020offline}. However, standard algorithms such as Fitted Q-Iteration (FQI~\citep{ernst2005tree}) can be over-optimistic in regions of the state--action space with low empirical support, leading to out-of-distribution recommendations. In pediatric ECMO, an unprecedented change in oxygenation or sweep gas may be clinically unacceptable. Imitation Learning (IL~\citep{osa2018algorithmic}) mitigates this risk by reproducing clinician behavior, but is limited by the demonstrator and cannot explicitly optimize long-term reward. This creates a performance--safety gap: a useful policy should improve estimated outcomes without departing recklessly from observed clinical practice.

\begin{figure}[t]
    \centering
    \begin{tabular}{@{}c@{\hspace{3pt}}c@{}}
        \includegraphics[width=0.416\linewidth]{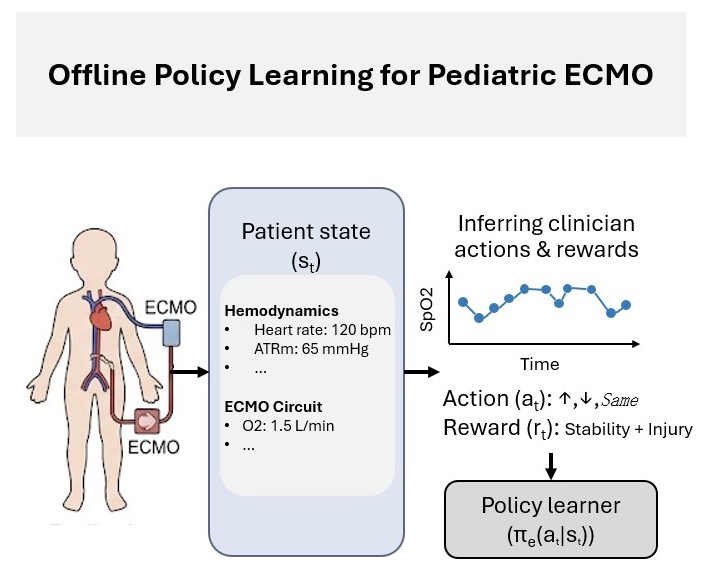}
        &
        \includegraphics[width=0.308\linewidth]{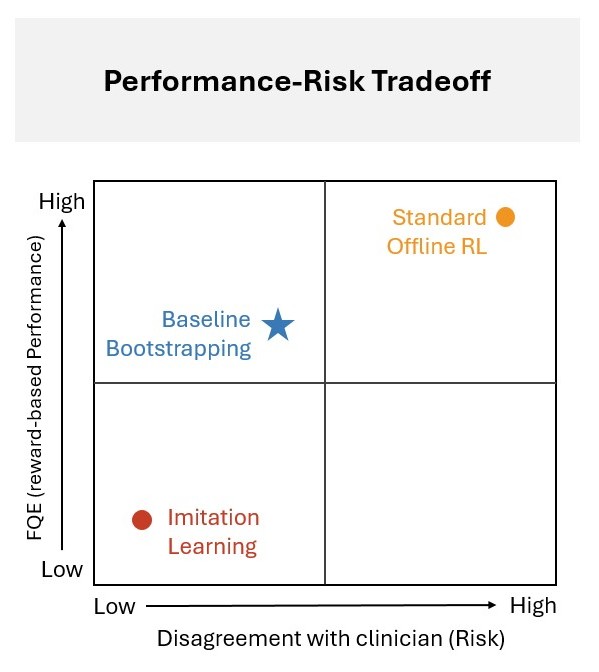}
        \\
        {\small (a)} & {\small (b)}
    \end{tabular}
    \caption{\textbf{Balancing Safety and Performance in Pediatric ECMO.} (a) We frame management of children on ECMO life support as an offline policy learning problem using logged clinician trajectories. (b) FQI attains high estimated value but deviates substantially from clinician behavior, whereas IL remains close to clinicians but has lower estimated value. FQI-SC seeks an intermediate operating point by restricting improvement to empirically supported actions.}
    \label{fig:overview}
\end{figure}
We bridge this gap with Support-Constrained Fitted Q-Iteration (FQI-SC), inspired by SPIBB~\citep{laroche2019safe,nadjahi2019safe}, which permits reward-driven optimization only where clinician data provide sufficient empirical support. We evaluate the resulting value--safety trade-off using a Balance Score $S(\pi)$, the harmonic mean of normalized FQE and behavioral conservatism. 
On 78 pediatric ECMO trajectories, we demonstrate that FQI-SC improves estimated value over IL while remaining substantially closer to clinician behavior than unconstrained FQI. Our contributions are:
\begin{itemize}
\item a \textbf{multi-objective reward signal} combining hourly physiological stability with neurological outcome;
    \item a \textbf{count-based support constraint} for offline RL over four clinically actionable ECMO controls;
    \item the \textbf{Performance-Safety Balance Score $S(\pi)$} for summarizing estimated value and behavioral conservatism;
    \item an empirical evaluation showing that the support threshold controls this trade-off.
    
    
\end{itemize}

\section{Background}
\paragraph{Pediatric ECMO and the Case for Decision Support}
Managing a patient on ECMO requires frequent, high-stakes adjustments to circuit and ventilator settings in response to 
rapidly changing physiology~\citep{moynihan2021decision}.
Building a decision-support tool to recommend actions for children on ECMO is challenging due to two factors.
First, codified knowledge about pediatric ECMO management is limited, in large part because ECMO is reserved for the most critically ill children. The 2022 ELSO Registry Report recorded only 2,394 pediatric ECMO runs in 2021 \citep{tonna2024elso}. Second, the gold standard for filling such gaps -- prospective hypothesis testing on patients -- is ethically infeasible when any new strategy would be tested on children whose lives depend on the current standard of care. Together, these constraints rule out online experimentation and force any learning approach to work from historical observational data alone. Therefore, pediatric ECMO decision-making falls within the framework of offline reinforcement learning.

\paragraph{Reinforcement Learning in Offline Settings}
We formalize treatment as a Markov Decision Process (MDP) 
$\mathcal{M}=(\mathcal{S},\mathcal{A},\mathcal{P},\mathcal{R},\gamma)$~\citep{sutton2018rl}. 
At time $t$, the agent observes state $s_t\in\mathcal{S}$, selects action $a_t\in\mathcal{A}$, receives reward $r_t$, and transitions to $s_{t+1}$. A policy $\pi(a\mid s)$ seeks to maximize expected discounted return $\mathbb{E}_\pi[\sum_t\gamma^t r_t]$, and the action-value function $Q^\pi(s,a)$ summarizes expected long-term return after taking action $a$ in state $s$. The optimal value satisfies the Bellman recursion
\(
Q^\star(s,a)=\mathbb{E}_{s'}[\mathcal{R}(s,a)+\gamma\max_{a'}Q^\star(s',a')].
\)
Offline RL~\citep{levine2020offline} replaces environment interaction with a fixed logged dataset $\mathcal{D}=\{(s_i,a_i,r_i,s'_i)\}$. Fitted Q-Iteration (FQI)~\citep{ernst2005tree} approximates $Q^\star$ by repeatedly fitting regression models to Bellman targets constructed from this dataset.

\paragraph{Distribution Shift and the Need for Safety}
Offline RL suffers from distribution shift: value functions may overestimate poorly supported actions, yielding potentially harmful recommendations~\citep{kumar2020conservative}. SPIBB~\citep{laroche2019safe} mitigates this risk by constraining policy improvement to sufficiently supported state--action pairs and reverting to a trusted baseline elsewhere. Under appropriate assumptions, SPIBB variants provide finite-sample improvement guarantees~\citep{nadjahi2019safe}. We adopt this support-based approach within FQI.

\paragraph{Imitation Learning as a Conservative Baseline}
Imitation learning (IL)~\citep{osa2018algorithmic} learns a policy from expert state--action demonstrations without optimizing long-term reward. We use IL as a conservative behavioral baseline for evaluating reward-driven policy improvement.


\paragraph{Off-Policy Evaluation}
A learned policy still cannot be tested prospectively on patients, so its value must be estimated from logged data. Off-policy evaluation (OPE)~\citep{precup2000eligibility,voloshin2019empirical} estimates the return of an evaluation policy $\pi_e$ from trajectories collected under another behavior policy. We use Fitted Q-Evaluation (FQE)~\citep{le2019fqe}, which fits an approximate $Q^{\pi_e}$ through a Bellman recursion specialized to the evaluation policy and averages predicted value over initial states. FQE generally avoids the extreme variance of trajectory-wise importance weighting but can be biased under model misspecification.

\paragraph{Clinical Reinforcement Learning}
The AI Clinician framework established a widely used template for offline clinical RL in sepsis~\citep{komorowski2018artificial}. Subsequent work has explored continuous treatment actions~\citep{huang2022reinforcement,wang2022learning}, richer EHR representations from clinical notes~\citep{lim2026moreclear}, and conservative value estimation for mechanical ventilation~\citep{kondrup2022towards}. ECMO-specific studies have learned probabilistic treatment rules~\citep{skinner2020preliminary}, applied fitted Q-learning to pediatric trajectories~\citep{ashtiani2020ecmo}, and developed offline RL for adult venovenous ECMO~\citep{song2026recmomender}. Most recently, a pediatric ECMO behavior-modeling study used a tabular foundation model to predict clinician actions~\citep{golivand2026imitation}. We build on that trajectory construction by adding reward-aware policy improvement with explicit empirical-support constraints.

\section{Support-Constrained Offline Policy Improvement}


\subsection{Problem Formulation}

We model the management of pediatric patients on ECMO as a Batch MDP $(\mathcal{S}, \mathcal{A}, \mathcal{P}, \mathcal{R}, \gamma)$. We treat each patient trajectory as a sequence of transitions $(s_t, a_t, r_t, s_{t+1})$, where $s_t \in \mathbb{R}^{F}$ represents the patient's physiological state at hour $t$, $a_t$ is the clinician action, $r_t$ is the scalar reward signal, and $F$ is the number of state features (Figure~\ref{fig:overview}(a)).

\paragraph{State and Action Spaces}
Our state and factored action formulation is inspired by the pediatric ECMO behavior-modeling framework of \citet{golivand2026imitation}, but we modify the action space to focus on variables that are directly controllable by clinicians. The state $s_t \in \mathbb{R}^{F}$ collects $F$ physician-selected physiological and
circuit features, and the global clinician action is decomposed into $\mathcal{K}$
independent control ``knobs'', each encoded as
$a_t^k \in \{\textsc{same}, \textsc{decrease}, \textsc{increase}\}$. This per-knob
factorization mirrors clinical decision-making and avoids the $3^\mathcal{K}$
combinatorial explosion of a joint action space. Defining each knob as a threshold-crossing event over a monitored physiological or support variable, rather than as a raw device setting,  ensures generalizability across different care teams, protocols, and device interfaces that may achieve the same physiological effect through different settings. On top of this action space, we learn an independent policy $\pi^k(a \mid s)$ for each knob.

\paragraph{Reward Shaping}
The reward function $r_t$ is designed to align the policy with two primary objectives: maintaining short-term \emph{physiological stability} and optimizing long-term \emph{clinical outcomes} (neurological injury). We define a state stability score, $\mathrm{Stab}(s)$, based on clinician-endorsed target ranges for $F$ hemodynamic and gas-exchange variables. Let $d_f(s)$ be the deviation of variable $f$ from its desired range at state $s$, defined such that $d_f(s)=0$ when the variable lies within the target range and increases as it moves farther outside. Let $w_f$ denote the clinician-assigned importance weight for variable $f$. The state stability score is:
\(
\mathrm{Stab}(s) = -\sum_{f=1}^{F} w_f \, d_f(s)
\)
where larger values correspond to physiologically superior states. 

\textbf{Neurologic-injury signal.}
We augment the hourly physiologic stability score with a sparse Neurologic Injury Score (NIS)-based component. Each NIS assessment is first classified as normal or abnormal according to a clinical injury threshold $\delta_{\mathrm{NIS}}$. Transitions between consecutive assessments are then categorized as recovery, sustained normal status, or new/persistent injury. Recovery receives a positive reward, sustained normal status receives no additional reward, and new or persistent injury receives a negative reward whose magnitude increases with worsening NIS.

In our implementation, we set $\delta_{\mathrm{NIS}}=8$, such that scores at or above 8 indicate neurologic injury. The first abnormal assessment receives $-5$, recovery from abnormal to normal receives $+5$, sustained normal status receives $0$, and new or persistent injury receives a severity-adjusted penalty ranging from $-5$ to $-10$ according to the change in NIS between consecutive assessments.

The reward is
\(
r_t =
\mathrm{Stab}(s_{t+1})-\mathrm{Stab}(s_t)
+
r_t^{\mathrm{NIS}},
\)
where $r_t^{\mathrm{NIS}}$ denotes the sparse neurologic-injury reward defined above.
By combining a change-in-stability term with the sparse neurologic-injury component, the reward encourages short-term physiological stability while remaining sensitive to the longer-term clinical trajectory. 

\paragraph{Objective}
For each head $k$, we seek a policy that maximizes the expected discounted return:
\(
J(\pi^k) = \mathbb{E}_{\pi^k} \left[ \sum_{t=0}^{T} \gamma^t r_t \right]
\)
where we use a discount factor of $\gamma = 0.99$, a standard choice for long-horizon RL problems.

\subsection{Fitted Q-Iteration on Continuous State}

Because the patient state is continuous and high-dimensional relative to the number of trajectories, we avoid tabular discretization of the state for value learning and instead use Fitted Q-Iteration (FQI) \citep{ernst2005tree} with gradient-boosted regressors. For each head $k$ and each action $a \in \{0,1,2\}$ we learn a separate XGBoost~\citep{chen2016xgboost} regressor $Q_a^k(s)$. We use XGBoost because it performs well on structured tabular data, captures nonlinear relationships, and is robust in moderate-data regimes.

Given the logged transitions $\mathcal{D} = \{(s_i, a_i^k, r_i, s'_i, \delta_i)\}_{i=1}^{|\mathcal{D}|}$, where $\delta_i$ indicates the terminal step of a trajectory, FQI proceeds iteratively. At iteration $m+1$, the target for transition $i$ is
\[
y_i^{(m+1)} \;=\; r_i + \gamma\,(1 - \delta_i)\,\max_{a' \in \mathcal{A}}\, Q_{a'}^{k,(m)}(s'_i),
\]
and trains the next iterate $Q_a^{k,(m+1)}$ on the subset of transitions where $a_i^k = a$. Thus, each regressor learns the action-value function associated with one specific action. We initialize the regressors with a reward-only fit (equivalent to a one-step return) and run $M{=}10$ Bellman iterations. The resulting greedy policy is
\(
\pi^k_{\mathrm{FQI}}(s) = \arg\max_a Q_a^k(s).
\)

\subsection{Behavior Model and Support Estimation}
\label{sec:support-constrained}

Plain FQI is known to be optimistic on out-of-distribution actions, which is a serious failure mode in a high-stakes ICU setting.  To mitigate this risk, we explicitly model the empirical support of the logged clinician policy and constrain the learned agent to remain within well-sampled regions of the state-action space. 

We estimate support with a coarse partition of the continuous state space built on the training fold only. After mean imputation and standardization (both fit on training data), we apply K-means to the training-fold states to obtain $K$ centroids, and assign each state to its nearest centroid, giving a cluster index $b(s_i) \in \{1,\dots,K\}$. The clustering is used only for support estimation, not as the input representation to FQI.
For each cluster $b \in \{1,\dots,K\}$ and each action $a$, we compute training-fold count
\(
N(b, a) \;=\; \sum_{i \in \text{train}} \mathbbm{1}[\,b(s_i) = b,\; a_i = a\,],
\)
These counts define a simple empirical count-based behavioral policy, which assigns action probabilities according to how frequently clinicians selected each action within the corresponding cluster:
\[
\hat{\pi}_b^k(a \mid s) \;=\; \frac{N(b(s), a) + \alpha}{\sum_{a'} N(b(s), a') + 3\alpha},\qquad \alpha = 10^{-3}.
\]
Thus, the count-based policy does not learn a parametric decision rule; instead, it represents the cluster-wise empirical action-frequency distribution observed in the logged data. We use this policy both as a behavioral anchor and as the fallback baseline for support-constrained FQI.

An action $a$ is considered \emph{supported} at state $s$ if $N(b(s), a) \geq N_\wedge$. We evaluate $N_\wedge \in \{5,10,15\}$. This thresholded count provides an operational definition of support: higher counts indicate that an action has been observed frequently enough in a region of the state space to justify reliable value estimation. As a result, increasing $N_\wedge$ makes the learned policy more conservative, as a larger proportion of state-action pairs are handled by falling back to the clinician's empirical distribution.

\subsection{Support-Constrained FQI}
\label{sec:spibb}

To learn a policy that cautiously improves upon the clinical baseline, we incorporate the support constraints of Section~\ref{sec:support-constrained} directly into value iteration, following \citet{laroche2019safe}.

For a given next-state $s'$, we partition the action space into a supported set $\mathcal{A}_{\mathrm{sup}}(s')$ and an unsupported (bootstrapped) set $\mathcal{B}(s') = \{a : N(b(s'),a) < N_\wedge\}$. Given the current Q-estimates and the baseline policy $\hat{\pi}_b^k(a\mid s')$, we calculate the projected next-state value as:
\[
V_{\mathrm{SPIBB}}(s') = \sum_{a \in \mathcal{B}(s')} \hat{\pi}_b^k(a\mid s')\, Q(s',a)  + \Bigg(\sum_{a \in \mathcal{A}_{\mathrm{sup}}(s')} \hat{\pi}_b^k(a\mid s')\Bigg)
\max_{a \in \mathcal{A}_{\mathrm{sup}}(s')} Q(s',a).
\]
If $\mathcal{A}_{\mathrm{sup}}(s')=\emptyset$, we fall back to the baseline expectation over the full action set. The Bellman target for each XGBoost regressor iterate is then:
\(
y_i^{\mathrm{SPIBB}} = r_i + \gamma (1-\delta_i) V_{\mathrm{SPIBB}}(s'_i).
\) At action selection, unsupported actions retain their baseline probabilities, while the total baseline probability mass of supported actions is assigned to the highest-$Q$ supported action; if no action is supported, we use the baseline policy. This formulation ensures that the policy is only permitted to redistribute probability mass among actions with sufficient empirical support.


\section{Cohort}

\paragraph{Cohort Selection}
We use retrospective trajectories from 78 pediatric ECMO patients without congenital heart disease (CHD), whose specialized management protocols fall outside our scope. The mean patient age is 4.66 years, and the mean trajectory length is 80 hours under the expert-defined observation windows.

\paragraph{Data Extraction}
For each patient we extract four categories of raw clinical data: hemodynamic measurements, ECMO circuit configuration parameters, ventilator settings and readings, and laboratory test results. The raw data are recorded at irregular intervals throughout each patient's ECMO course. We adopt the preprocessing protocol of \citet{golivand2026imitation}, developed for pediatric ECMO imitation learning: each trajectory is discretized into hourly windows, values within each window are aggregated by the mean, and the physician-selected physiological features are retained from the full set of recorded variables.
These variables span hemodynamics (mean arterial pressure, heart rate, SpO\textsubscript{2}, and cerebral oximetry rSO\textsubscript{2}-1), ECMO-circuit settings (blood flow, sweep CO\textsubscript{2}/O\textsubscript{2} flow, volume sensor, and FiO\textsubscript{2}), ventilator parameters (mean airway pressure, PEEP, and tidal volume), and laboratory values (arterial PO\textsubscript{2}, PCO\textsubscript{2}, pH, base excess, lactate, ionized calcium, total CO\textsubscript{2}, and INR); the complete list with units follows \citet{golivand2026imitation}, with the minor difference that rSO\textsubscript{2}-2 and etCO\textsubscript{2} are excluded following clinician review.

\paragraph{Trajectory truncation.}
Trajectory duration was restricted using an expert-guided definition of the clinically relevant observation window. For patients whose NIS assessments remained within the normal range, we retained the first 48 hours after ECMO initiation, focusing the analysis on the early phase of ECMO management. For patients with neurologic injury that persisted or worsened, the trajectory was retained through the final NIS assessment, so that the observed course leading to the most recent documented neurological outcome was preserved.

\subsection{Feature Construction}

\paragraph{State representation.} Inspired by \citet{golivand2026imitation}, at each time step $t$, the state representation includes, for each physician-selected variable, both (i) the mean value within hour $t$ and (ii) the delta relative to hour $t-1$, capturing the current physiological setpoint and a short-term trend. Heart rate (HR) and mean arterial pressure (ARTm) are age-normalized, since the clinically meaningful range of these values varies substantially across pediatric age groups. The type of ECMO (VV or VA) together with a binary \texttt{on-ecmo} indicator are also added to the features. The resulting vector is used as input to the FQI Q-function, the FQE Q-function, and the imitation-learning classifiers.

\paragraph{Action labels.} We define four clinically actionable control variables: ECMO blood flow, ECMO oxygen flow, ventilator oxygen concentration, and ECMO FiO$_2$. For each control variable $k$, the hourly change relative to the previous window is compared against a physician-defined threshold $\delta_k$: the label is \textsc{increase} or \textsc{decrease} when the signed change exceeds $\pm\delta_k$, and \textsc{same} otherwise. The corresponding change thresholds are 10\,mL/kg/min for ECMO blood flow, 0.1\,L/min for ECMO O$_2$ flow, 5 percentage points for ventilator oxygen concentration, and 0.05 for ECMO FiO$_2$. These thresholds distinguish clinically meaningful interventions from sensor noise and physiological fluctuation, yielding a 4-dimensional discrete action vector $a_t = (a_t^1, \dots, a_t^4)$ at each hour.

\paragraph{Reward signal.} Our reward signal consists of two components: (i) an hourly physiological stability component and (ii) a sparse neurologic-injury component based on transitions between NIS assessments.
To define the stability term, we asked physicians to specify desirable ranges for major hemodynamic, gas-exchange, and perfusion variables, together with an importance weight for each clinical category. Table~\ref{tab:reward_targets} summarizes both the target ranges and the corresponding clinician-provided importance weights.

\begin{table}[t]
  \centering
  \caption{Clinician-defined target ranges and importance weights used in the
  stability component of the reward. Weights are assigned per clinical category.}
  \label{tab:reward_targets}
  \small
  \setlength{\tabcolsep}{5pt}
  \begin{tabular}{lllc}
  \toprule
  \textbf{Variable} & \textbf{Target / Rule} & \textbf{Unit} & \textbf{Weight} \\
  \midrule
  BP & (0--1) & age-norm. & 2.0 \\
  HR & (0--1) & age-norm. & 1.5 \\
  SpO$_2$ & 92--100 & \% & 1.5 \\
  PaO$_2$ & 80--150 & mmHg & 1.5 \\
  PaCO$_2$ & 35--45 & mmHg & 2.0 \\
  Arterial pH & 7.30--7.45 & -- & 1.5 \\
  rSO$_2$-1 & 60--95 & \% & 1.5 \\
  Lactate & low $<$ 2, high $>$ 10 & mmol/L & 2.0 \\
  \bottomrule
  \end{tabular}
\end{table}

\section{Results on Real Data} 
We study two questions.  
    \textbf{Q1.} Can a policy be learned safely from limited offline pediatric ECMO data?
    \textbf{Q2.} Is there a controllable way to bias the policy toward safety or toward reward maximization?

\subsection{Experimental Setup}
We use 3-fold GroupKFold cross-validation by patient so that no patient appears in both training and validation folds. Support is estimated on each training fold using the same K-means partition with $K=100$ clusters described in Section~\ref{sec:support-constrained}. All XGBoost regressors used for FQI and FQE were trained with the same fixed configuration across folds, heads, and policy variants, using 30 estimators, maximum depth 4 and learning rate 0.1.

\subsection{Baselines}
We compare our support-constrained policy (FQI-SC) with plain FQI, and an imitation learning baseline.
Plain FQI uses the logged transitions to optimize reward but does not explicitly model or imitate the clinician behavior policy, whereas IL directly learns to mimic clinician actions without using the reward.

For the imitation learning baseline, we adopt the per-knob hierarchical two-stage classifier of \citet{golivand2026imitation}: for each knob, a first binary model predicts whether the knob changes and a second predicts the direction given a change, combined into a 3-way distribution over $\{\textsc{same}, \textsc{decrease}, \textsc{increase}\}$. This decomposition addresses the heavy class imbalance toward \textsc{same}. Both stages use TabPFN \citep{tabpfn,golivand2026imitation}, a transformer-based prior-data fitted network for tabular classification; that study evaluated TabPFN against XGBoost and MLP on the same setting and found stronger generalization and substantially better calibration. We use this model only as a behavioral-cloning comparison baseline. The behavioral anchor used internally by FQI-SC is instead the count-based empirical policy $\hat{\pi}_b^k$ defined in Section 3.

\subsection{Evaluation Metrics}
Because candidate policies cannot be deployed on patients, evaluation is performed off-policy on held-out trajectories. We report FQE for estimated policy value, disagreement with logged clinician actions for behavioral conservatism, and their combined Performance-Safety Balance Score $S(\pi)$. FQE uses 10 Bellman iterations and policy smoothing $\epsilon=0.02$.

\paragraph{Fitted Q-Evaluation (FQE).} 
For each candidate policy $\pi_e$ we run a separate Bellman recursion on the training fold,
\(
y_i^{(m)} = r_i + \gamma (1 - \delta_i)\sum_{a'} \pi_e(a' \mid s'_i)\, \hat{Q}_{a'}^{(m-1)}(s'_i),
\)
where $\delta_i \in \{0,1\}$ is the terminal-state indicator and
$\hat{Q}_{a'}^{(m-1)}$ denotes the previous FQE iterate. Each regressor
$\hat{Q}_a^{(m)}$ is then re-fit from scratch on the subset of training transitions
for which $a_i = a$, using $y_i^{(m)}$ as the supervised target. After the final iteration, we report
\[
\hat{V}_{\pi_e}^{\mathrm{FQE}}
=
\mathbb{E}_{s_0}\!\left[\sum_a \pi_e(a \mid s_0)\, \hat{Q}_a(s_0)\right],
\] the average initial-state value on held-out patient trajectories, where $\hat{Q}_a$ is the final learned action-value function for action $a$ under $\pi_e$.



\paragraph{Disagreement rate.}
For each policy, we also report the fraction of held-out states for which the policy's hard action prediction differs from the logged clinician action. Let
\(
\hat{a}_i^{\pi} = \arg\max_{a} \pi(a \mid s_i)
\)
denote the hard action predicted by policy $\pi$ at held-out state $s_i$, and let $a_i^{\mathrm{log}}$ denote the corresponding logged clinician action. The disagreement rate is then
\(
d(\pi)
=
\frac{1}{N_{\mathrm{val}}}
\sum_{i=1}^{N_{\mathrm{val}}}
\mathbbm{1}\!\left[\hat{a}_i^{\pi} \neq a_i^{\mathrm{log}}\right].
\)
Lower disagreement therefore indicates a more behaviorally conservative policy.

\paragraph{Performance-Safety Balance Score.} To summarize the trade-off between estimated value and behavioral conservatism in a single quantity, we define a combined score based on normalized FQE and normalized disagreement rate. For a policy $\pi$, let
\[
V_{\mathrm{norm}}(\pi)
=
\frac{\widehat{V}^{\mathrm{FQE}}(\pi)-V_{\min}}{V_{\max}-V_{\min}},
\quad
A_{\mathrm{norm}}(\pi)
=
\frac{d_{\max}-d(\pi)}{d_{\max}-d_{\min}},
\]
where $\widehat{V}^{\mathrm{FQE}}(\pi)$ is the FQE estimate and $d(\pi)$ is the disagreement rate. $V_{\min}$, $V_{\max}$, $d_{\min}$, and $d_{\max}$ are selected over the full set of evaluated policy--head--fold combinations, using FQE for the value bounds and disagreement for the behavioral bounds. We then define
\[
S(\pi)
=
\frac{2\,V_{\mathrm{norm}}(\pi)\,A_{\mathrm{norm}}(\pi)}
{V_{\mathrm{norm}}(\pi)+A_{\mathrm{norm}}(\pi)}.
\]
A higher $S(\pi)$ indicates a better balance between estimated long-term value and closeness to historical clinician behavior. By construction, this score favors policies that perform reasonably well on both axes and penalizes policies that achieve a strong result on only one.
\begin{figure}[t]
    \centering
    \includegraphics[width=\linewidth]{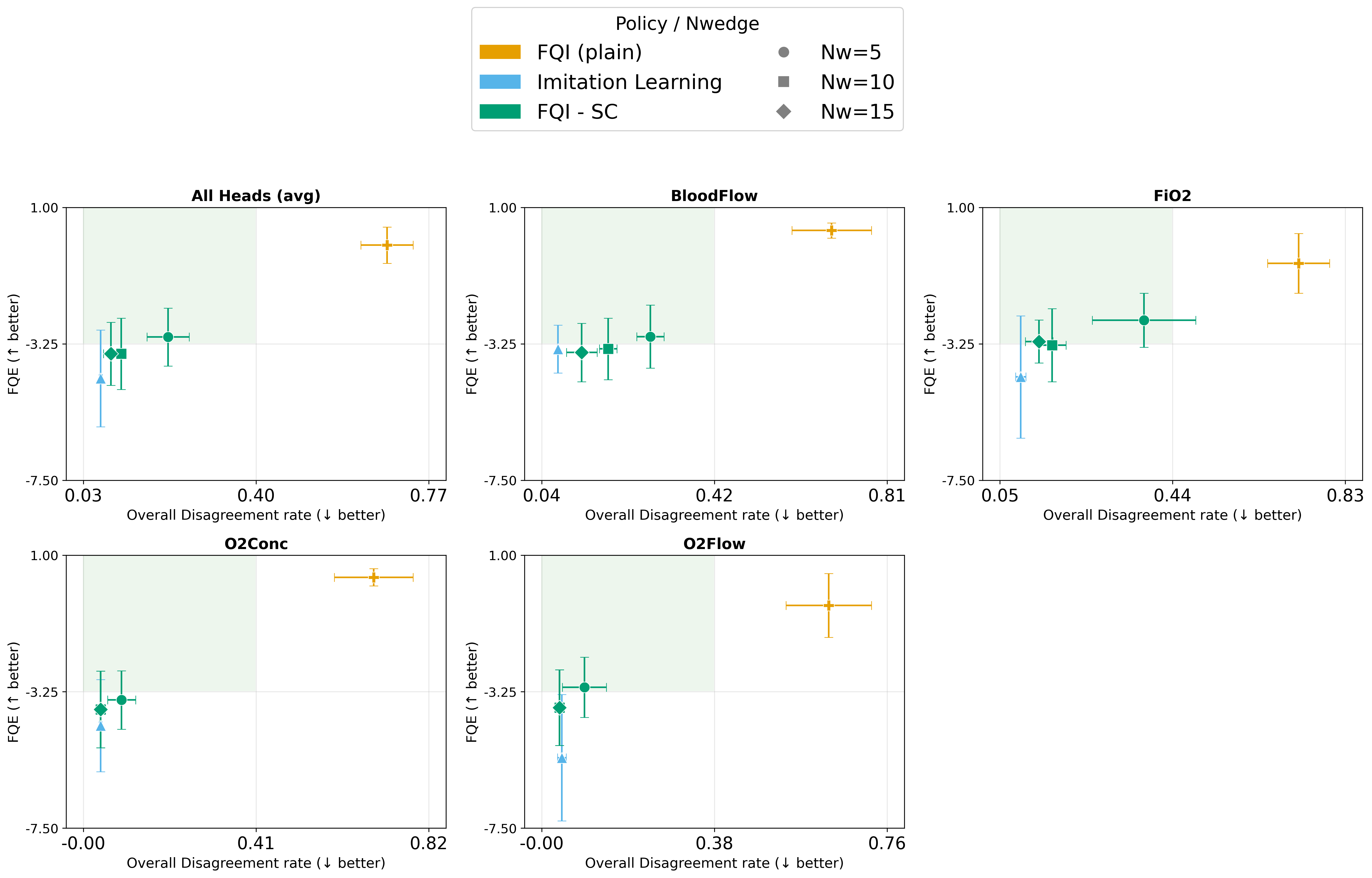}
    
    \caption{Per-head and all-head trade-off between estimated value and behavioral conservatism. Each point shows a policy's mean FQE and disagreement rate, with error bars denoting variability across folds. Plain FQI occupies the high-value but high-disagreement regime, imitation learning occupies the low-disagreement but lower-value regime, and support-constrained FQI (FQI-SC) lies between these extremes. Different marker shapes indicate the support threshold $N_\wedge \in \{5,10,15\}$, and the shaded upper-left region highlights the desirable operating regime of higher FQE and lower disagreement.}
    \label{fig:fqeVSdisagreement}
\end{figure}

\subsection{Q1: Can a policy be learned safely from limited offline data?}

To evaluate the extent to which offline reinforcement learning can discover treatment strategies, we evaluate the estimated value and safety scores across the three methods: plain FQI, TabPFN-based Imitation Learning, and FQI-SC. Figure~\ref{fig:fqeVSdisagreement} visually summarizes the trade-off between estimated value and behavioral conservatism across heads and support thresholds.

As shown in Table~\ref{tab:overall_tradeoff}, 
FQI achieves the highest estimated value (\(-0.18\pm0.28\)) but the largest disagreement (\(0.68\)), whereas IL has the lowest disagreement (\(0.06\)) but lower estimated value (\(-4.32\pm0.56\)). FQI-SC provides a middle ground: at \(N_\wedge=5\), FQE is \(-3.04\pm0.20\), disagreement is \(0.20\), and the Balance Score is highest among the compared settings (\(S=0.65\)). This corresponds to a 71\% reduction in disagreement relative to FQI and a 1.28-unit FQE improvement over IL.



\begin{table}[t]
\centering
\caption{Overall performance across all heads. Higher FQE indicates better estimated long-term reward, while lower disagreement indicates closer adherence to historical clinician behavior. The score $S$ summarizes the balance between these two objectives. Green cells indicate more desirable values, red cells indicate less desirable values. FQE values are reported as mean $\pm$ standard deviation across head--fold evaluations.}
\label{tab:overall_tradeoff}
\small
\setlength{\tabcolsep}{5pt}
\begin{tabular}{lccc}
\toprule
\textbf{Policy} & \textbf{Avg FQE $\uparrow$} & \textbf{Disagreement $\downarrow$} & \textbf{Score $S \uparrow$} \\
\midrule
FQI & \cellcolor{green!40} \textbf{-0.18{\tiny $\pm$0.28}} & \cellcolor{red!22} 0.68 & \cellcolor{green!15} 0.47 \\
Imitation Learning & \cellcolor{red!22} -4.32{\tiny $\pm$0.56} & \cellcolor{green!50} \textbf{0.06} & \cellcolor{green!10} 0.50 \\
FQI--SC $N_\wedge=5$ & \cellcolor{green!20} -3.04{\tiny $\pm$0.20} & \cellcolor{green!20} 0.20 & \cellcolor{green!52} \textbf{0.65} \\
\bottomrule
\end{tabular}
\end{table}

\subsection{Q2: Is there a controllable way to bias the policy toward safety or toward reward maximization? }

\begin{table}[t]
\centering
\caption{Effect of the support threshold $N_\wedge$ on support-constrained FQI (FQI-SC) across all heads. The score $S$ summarizes the balance between these two objectives using the same normalization and harmonic-mean definition as in Table~\ref{tab:overall_tradeoff}.}
\label{tab:nwedge_tradeoff}
\small
\setlength{\tabcolsep}{5pt}
\begin{tabular}{lccc}
\toprule
\textbf{$N_\wedge$} & \textbf{Avg FQE $\uparrow$} & \textbf{Disagreement $\downarrow$} & \textbf{Score $S \uparrow$} \\
\midrule
15 & \cellcolor{green!10} -3.56{\tiny $\pm$0.14} & \cellcolor{green!50} \textbf{0.09} & \cellcolor{green!20} 0.63 \\
10 & \cellcolor{red!10} -3.56{\tiny $\pm$0.13} & \cellcolor{green!12} 0.11 & \cellcolor{green!10} 0.63 \\
5  & \cellcolor{green!20} \textbf{-3.04{\tiny $\pm$0.20}} & \cellcolor{red!10} 0.20 & \cellcolor{green!52} \textbf{0.65} \\
\bottomrule
\end{tabular}
\end{table}

To evaluate whether the policies learned by FQI-SC can be biased towards safety and reward maximization, we evaluate three variants with $N_\wedge$ set to 5, 10, and 15. Table~\ref{tab:nwedge_tradeoff} shows that \(N_\wedge\) controls this trade-off. Increasing it from 5 to 10 reduces disagreement from \(0.20\) to \(0.11\), while FQE changes from \(-3.04\) to \(-3.56\). At \(N_\wedge=15\), disagreement falls further to \(0.09\) while mean FQE remains \(-3.56\). The corresponding Balance Scores are \(0.65\), \(0.63\), and \(0.63\). Thus lower \(N_\wedge\) favors estimated value, whereas higher thresholds produce more behaviorally conservative policies.


\section{Conclusion}
We developed FQI-SC for cautious offline policy improvement in pediatric ECMO. By grounding optimization in empirical support, FQI-SC reduces the large clinician disagreement produced by unconstrained FQI while retaining higher estimated value than imitation learning. Among the tested thresholds, $N_\wedge=5$ has the highest Balance Score ($0.65$), improving FQE by 1.28 units over IL and reducing disagreement by 71\% relative to FQI.

The study remains limited by its modest retrospective cohort, possible unmeasured confounding, and reliance on model-based off-policy evaluation rather than prospective outcomes. These factors limit both causal interpretation and generalizability. External multi-center validation and direct clinician evaluation are therefore necessary before clinical use. Future work could also investigate learned reward models, including inverse reinforcement learning~\citep{abbeel2004irl}, to reduce dependence on expert-specified reward components.

\section{Responsible-Use Statement}
This work is intended for research and clinical decision-support development, not for autonomous treatment recommendation or direct clinical deployment. All results are based on retrospective observational data, and neither FQE nor agreement with historical clinician behavior establishes causal treatment benefit or clinical safety. The learned policies may reflect unmeasured confounding, biases in historical practice, and limitations of the available patient cohort, while the expert-specified reward targets and action thresholds represent only a subset of relevant clinical knowledge. Before any clinical use, the approach would require external and prospective validation, evaluation across institutions and patient subgroups, and direct review of recommendations by ECMO clinicians.

\begin{ack}
We acknowledge support from NIH award R01NS133142.
\end{ack}

\bibliographystyle{plainnat}
\bibliography{sample}

\end{document}